\documentclass[10pt,twocolumn,letterpaper]{article}

\usepackage{ijcb}
\usepackage{times}
\usepackage{epsfig}
\usepackage{graphicx}
\usepackage{amsmath}
\usepackage{amssymb}
\usepackage[norule,symbol,perpage]{footmisc}

\usepackage{hhline}

\ijcbfinalcopy 

\ifijcbfinal\fi

\makeatletter
\def\ps@IEEEtitlepagestyle{
\def\@oddfoot{\mycopyrightnotice}
\def\@evenfoot{}
}
\def\mycopyrightnotice{
{\hfill \footnotesize 978-1-7281-9186-7/20/\$31.00 \copyright 2020 IEEE\hfill}
}
\makeatother

\begin{document}

\title{Cross-Spectral Periocular Recognition with Conditional Adversarial Networks}

\author{Kevin Hernandez-Diaz, Fernando Alonso-Fernandez, Josef Bigun\\
School of Information Technology, Halmstad University, Sweden\\
{\tt\small kevin.hernandez-diaz@hh.se, feralo@hh.se, josef.bigun@hh.se}
}

\maketitle
\thispagestyle{empty}

\begin{abstract}
%
This work addresses the challenge of comparing periocular images captured in different spectra,
which is known to produce significant drops in performance in comparison to operating in the same spectrum.
We propose the use of Conditional Generative Adversarial Networks,
trained to convert periocular images between visible and near-infrared spectra,
so that biometric verification is carried out in the same spectrum.
The proposed setup allows the use of existing feature methods
typically optimized to operate in a single spectrum. 
Recognition experiments are done using a number of 
off-the-shelf periocular comparators based both on
hand-crafted features and CNN descriptors.
Using the
Hong Kong Polytechnic University Cross-Spectral Iris Images Database
(PolyU) as benchmark dataset,
our experiments show that cross-spectral performance is substantially
improved if both images are converted to the same spectrum,
in comparison to matching features extracted from images in different spectra.
In addition to this, we fine-tune a CNN based on the ResNet50 architecture,
obtaining a cross-spectral periocular performance of EER=1\%, and
GAR$>$99\% @ FAR=1\%, which is comparable to the state-of-the-art with the PolyU database.
%

\end{abstract}

\let\thefootnote\relax\footnotetext{\mycopyrightnotice}

\section{Introduction}

Periocular recognition makes use of the region surrounding the eye 
to determine the identity of a person \cite{[Alonso16]}.
This region can be used for recognition purposes
in a wide range of unconstrained environments
where the face may be occluded (e.g. due to masks or uncooperative subjects), 
or the iris has not sufficient resolution \cite{[Nigam15]}.
In addition to requiring a more relaxed acquisition than face or iris,
the periocular modality
is more resistant to covariates such as
face expression \cite{[Park11]},
image blur \cite{[Miller10]}, or
aging \cite{[Juefei-Xu11]}.
It can be also used to reliably predict for example
gender or ethnicity \cite{[Alonso16]},
or to estimate face expression \cite{[Alonso18_perioc_expression]}.
Given than the periocular area appear in iris and face images,
it can be easily obtained with existing face and iris setups as well.

In spite of the advances in periocular biometrics,
cross-spectral operation remains a challenge.
%
%
A significant performance degradation is usually observed
when comparing periocular images acquired in different spectra \cite{[Jillela14],[Sharma14],[Ramaiah16],[Hernandez19]}.
Iris images are usually captured in the near-infrared (NIR) spectrum,
with many large-scale applications in place such as national ID programs \cite{[Nalla17]}.
%
In parallel, the use of face images in the visible (VIS) spectrum is
booming in applications involving social networks, smartphone devices,
or surveillance cameras.
Therefore, in many practical applications, 
it may be the case that the registration and test images are not
acquired in the same spectrum \cite{[Jillela14]}. 

\begin{figure}[b]
\centering
        \includegraphics[width=0.32\textwidth]{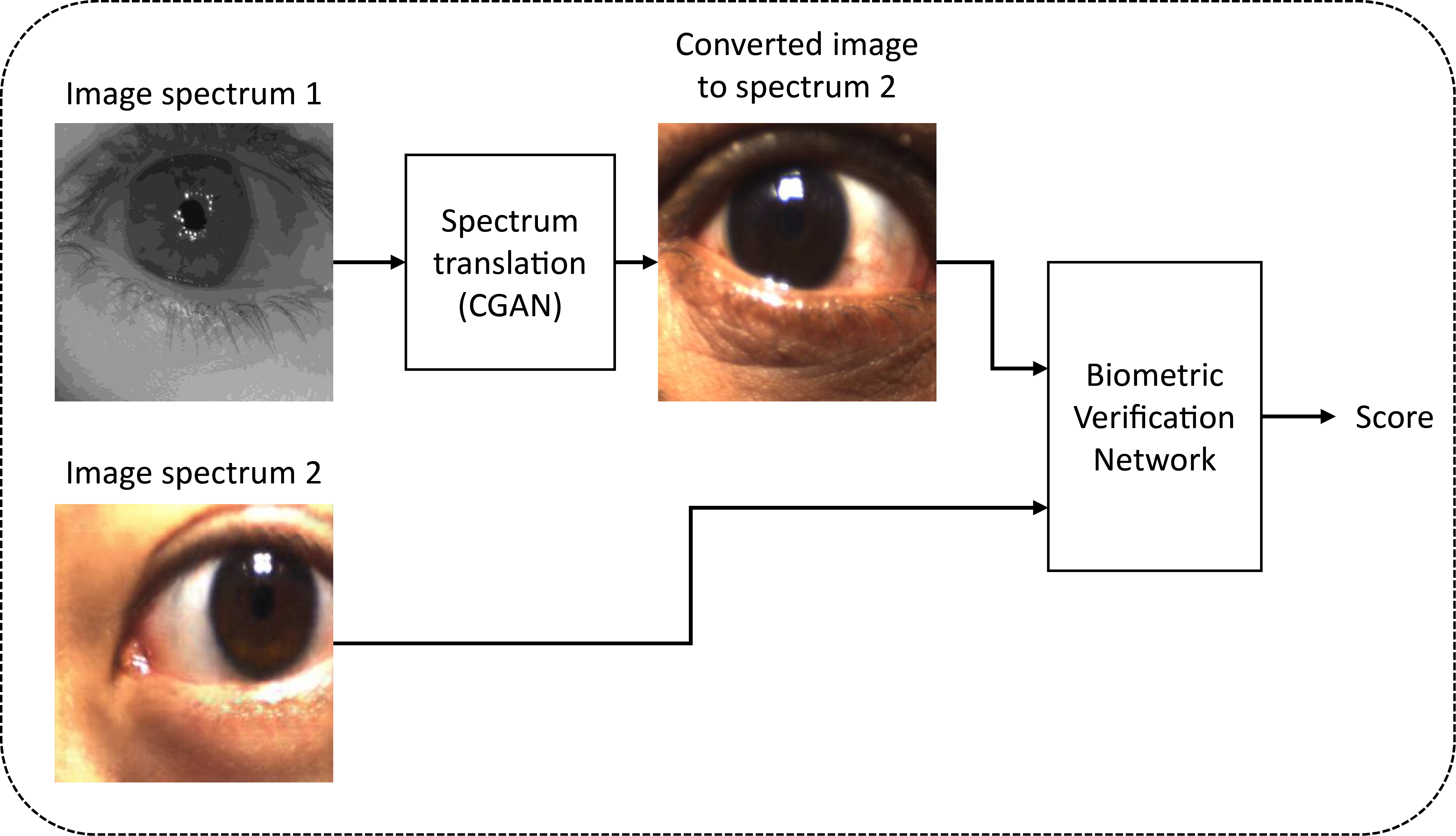}
\caption{Our cross-spectral periocular recognition framework.}
\label{fig:system_model}
\end{figure}

Accordingly, this paper addresses the challenge of 
cross-spectral periocular recognition.
%
%
This has been studied in a number of works.
Operational evaluation is usually done by reporting the Genuine Acceptance Rate 
(GAR, or proportion or genuine users correctly accepted) 
for a given False Acceptance Rate 
(FAR, or proportion of unauthorized persons accepted by the system). 
A false acceptance typically is considered the most serious of biometric errors, as it gives unauthorized users access to systems that expressly are trying to keep them out.
Therefore, operational points of FAR=1\%, 0.1\% or even smaller are typically employed. 
Another metric is the Equal Error Rate (EER), which corresponds to the operational point when the proportion of false acceptances is equal to the proportion of false rejections.
The work \cite{[Jillela14]} proposed to compare
the ocular region cropped from VIS face images against NIR iris
images. 
They employed three feature descriptors, namely Local Binary Patterns (LBP),
Normalized Gradient Correlation (NGC), and
Joint Database Sparse Representation (JDSR).
Using a self-captured database with 1358 images of the left eye from 704 subjects,
they report a
cross-spectral performance of EER=23\% by score-level
fusion of the three experts.
The authors in \cite{[Sharma14]} presented the
IIITD Multispectral database (IIITD-IMP), with 1240
VIS, NIR and Night Vision images from 62 subjects.
To cope with cross-spectral comparisons, 
they employed Neural Networks (NNs) to learn the variabilities
caused by each pair of spectra. 
They used a Pyramid
of Histogram of Oriented Gradients (PHOG) as input of the Neural Networks. 
They reported results for each eye separately, and
for the combination of both eyes, with 
cross-spectral performance between GAR=38-64\% at FAR=1\% (best of the two eyes)
and GAR=47-72\% (combining the two eyes).
The use of off-the-shelf Convolutional Neural Networks (CNNs)
as feature extraction method for NIR-VIS comparison
was recently proposed in \cite{[Hernandez19]}.
Here, the authors identified the layer of the ResNet101
network (pre-trained on ImageNet) 
that provides the best performance on each spectrum.
Then, they trained a NN that uses as input 
the feature vector of the best respective layers. 
Using the IIITD-IMP database, 
they reported results considering the left and right eyes of a person as different users (effectively duplicating the number of classes). 
The obtained cross-spectral accuracy was
EER=5-10\% and GAR=81-88\% at FAR=1\%,
outperforming any previous study with this database.

The same database employed in the present paper,
the Hong Kong Polytechnic University Cross-Spectral Iris Images Database (PolyU) \cite{[Nalla17]},
has been used in a number of studies with the periocular and iris modalities, 
or a fusion of both.
It has 12540 images from 209 subjects in NIR and VIS spectra.
A comparative summary of works using this database, including the
present paper, 
can be found later in the experimental section (Table~\ref{tab:otherworks-results}).
It must be highlighted that the majority of studies employ a 
subset of the database which corresponds to subjects 
whose irises can be properly segmented.
The authors in \cite{[Ramaiah16]} employed the IIITD-IMP and PolyU databases,
with each eye considered a different user.
To carry out VIS-NIR comparisons,
they used Markov Random Fields combined with two 
variants of local binary patterns (LBP), namely, FPLBP
(Four-Patch LBP) and TPLBP (Three-Patch LBP).
They reported a cross-spectral periocular 
GAR at 0.1\% FAR
of 16-18\% (IIITD-IMP) and 45-73\% (PolyU).
By fusion with the iris modality using Log-Gabor descriptors, 
they further improved the results up to GAR=74-84\% with PolyU.
The IIITD-IMP, PolyU and CrossEyed databases 
(the latter with 3840 images in NIR and VIS spectra from 
120 subjects \cite{[sequeira16crosseyed]}) were used in \cite{[Behera17]}.
To normalize differences in illumination between NIR and VIS images,
they applied Difference of Gaussian (DoG) filtering.
The descriptors employed included
LBP and HOG.
They reported results for each eye separately, and
for the combination of both.
The IIITD-IMP database gave the worst results, with
a cross-spectral EER of 45\% and a GAR at 0.1\% FAR of only 25\% 
(two eyes combined).
The reported accuracy with the other databases is better, ranging between 10-14\% (EER) and 83-89\% (GAR).

Another works report results only with the iris modality
of PolyU \cite{[Nalla17],[Wang19]}.
These are also included here 
for comparative purposes.
The authors of \cite{[Nalla17]} investigated two approaches
to cope with cross-spectrum recognition.
First, they proposed a classification framework based on
Naive-Bayes Nearest-Neighbor (NNBN).
In the second approach, they developed a iris texture synthesis
framework using multiscale Markov random fields (MRF),
where they estimated VIS iris patterns from the synthesis 
of iris patches in NIR images. 
The reported iris accuracy with PolyU is EER=24-27\% and 
GAR=59-62\% at 0.1\% FAR.
In \cite{[Wang19]}, they incorporated
Supervised Discrete Hashing (SDH) for compression and classification
of the features learned from an own trained CNN with softmax cross-entropy loss.
The SDH framework was also incorporated into other existing CNN
architectures (VGG16 and ResNet50).
The reported iris accuracy using PolyU is further reduced in comparison to the previous study
to EER=5-7\% and GAR=77-87\% at 0.1\% FAR 
using a subset of 280 classes,
and to EER=12\% and GAR=57\% over the entire dataset.
The entire dataset is also used in \cite{[Zanlorensi20deepXspectralOcular]} (together with the CrossEyed database), where the authors fine-tuned a VGG16 and a ResNet50 model pre-trained for face recognition \cite{[Parkhi15],[Cao18vggface2]}. The networks were first trained for biometric identification, 
using images from both spectra together. 
To carry out verification experiments, the vector before the cross-entropy layer were used as feature representation, and the cosine distance was used as metric to match two given samples.
The reported cross-spectral periocular accuracy is EER=1.8$\pm$0.21\% (VGG16) and 0.78$\pm$0.09\% (ResNet50).
They also reported results of the iris modality (being worse than the periocular modality), and of the fusion of both modalities, pushing the error rates further down. 
%


\begin{figure}[t]
\centering
        \includegraphics[width=0.42\textwidth]{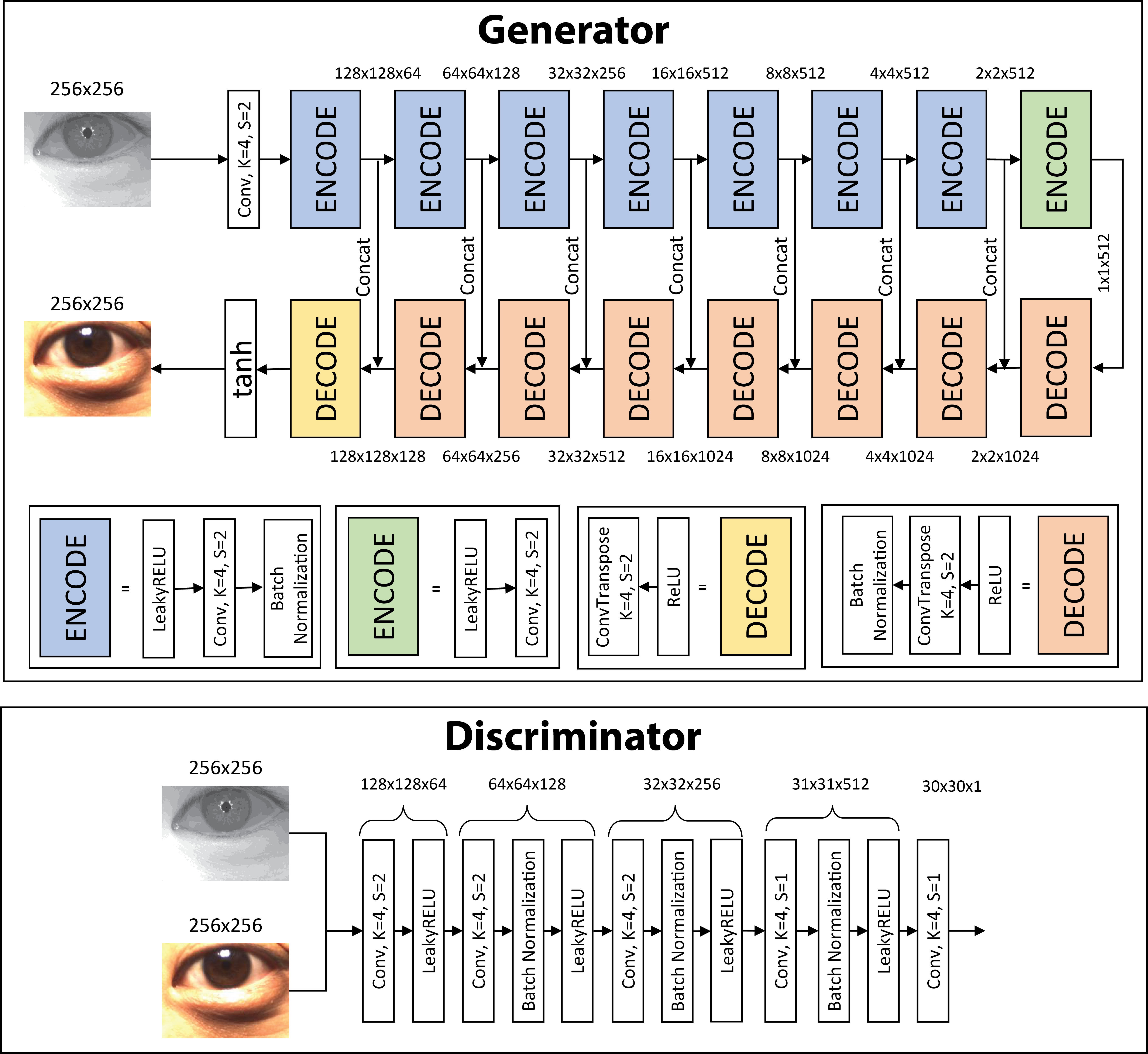}
\caption{Generator and discriminator models.}
\label{fig:gan_pix2pix_models}
\end{figure}

In this work, we address the challenge of periocular
cross-spectrum recognition 
using a image translation technique based on
Conditional Generative Adversarial Networks (CGANs) \cite{[mirza2014cgans],[isola16pix2pix]}.
In this approach (Figure~\ref{fig:system_model}),
the image in one spectrum is converted to the other spectrum, so that verification is carried out in the same spectral domain. 
We employ the PolyU database in our setup. 
The proposed approach allows to use existing biometric comparators without further modification.
For verification experiments, we employ the traditional 
Histogram of Oriented Gradients (HOG),
Local Binary Patterns (LBP) and 
Scale-Invariant Feature Transform (SIFT) key-points, 
used as baseline in many periocular studies \cite{[Alonso16]}.
Inspired by previous works in ocular biometrics \cite{[Nguyen18],[Hernandez18],[Hernandez19]}, 
we also use three off-the-shelf CNNs pre-trained on the ImageNet database (Resnet101 \cite{[He16]}, Densenet201 \cite{[Huang17]},  MobileNet v2 \cite{[Sandler18mobilenetv2]}),
and two CNNs pre-trained to classify faces (VGGFace \cite{[Parkhi15]}, ResNet50ft \cite{[Cao18vggface2]}).
These networks have proven to be successful in recognition tasks apart from the task for which they were designed, including cross-spectral periocular recognition \cite{[Hernandez19],[Alonso20inffus]}.
In our experiments, we observe that the cross-spectral performance of these off-the-shelf descriptors improve substantially if the images are converted to the same spectrum, validating the proposed CGAN-based spectrum translation technique.
To pursue even lower cross-spectral error rates, we fine-tune a ResNet50 architecture pre-trained on the ImageNet database (which contains about 1M images from 1000 classes). The error rates obtained are competitive with the state-of-the-art, given by recent study \cite{[Zanlorensi20deepXspectralOcular]}. The latter employs ResNet50 too, but using as base the network of \cite{[Cao18vggface2]}, which is fine-tuned first for face recognition with two databases that are much bigger than ImageNet (MS-Celeb-1M, with 10M images from 100K people, and VGGFace2 dataset, with 3.1M images from 8.6K people).
From this viewpoint, we are capable of obtaining comparable performance with a much smaller training budget.


The rest of the paper is organized as follows. In Section 2, 
we describe the spectrum translation method based on 
Conditional Generative Adversarial Networks.
Section 3 presents 
our experimental framework, including
the database, the 
protocol employed for cross-spectral periocular recognition, the periocular comparators, and the results.
Finally, conclusions are given in Section 4.

\begin{figure}[t]
\centering
        \includegraphics[width=0.42\textwidth]{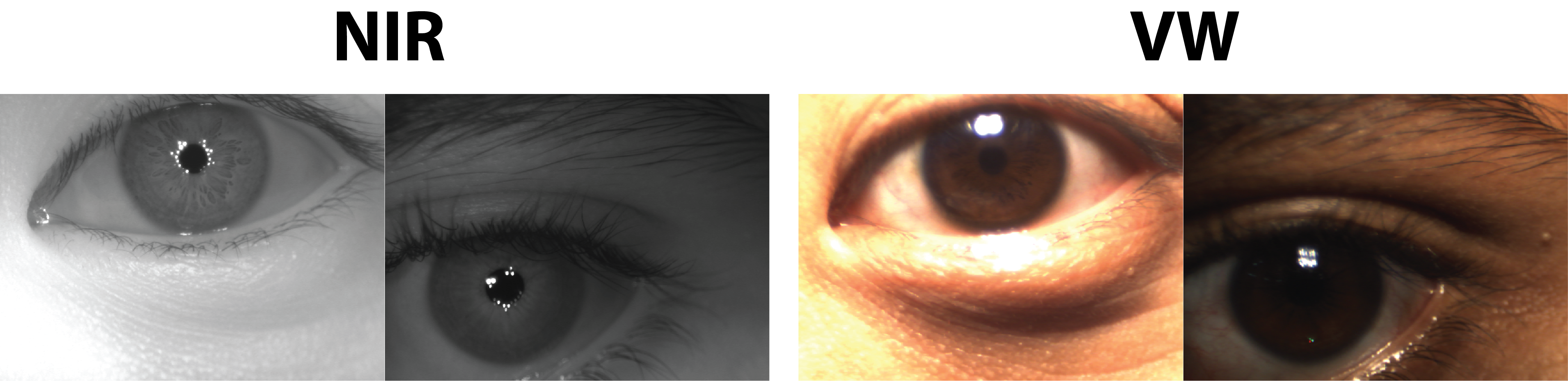}
\caption{Images from the PolyU database.}
\label{fig:sample_images_polyu}
\end{figure}

\section{Spectrum Translation}
\label{sect:spectrum_translation}

This section describes the spectrum translation algorithm employed. 
Our aim is to make that the images look as if they were captured in the target spectrum.
To accomplish this, we employ Conditional Generative Adversarial Networks (CGANs) \cite{[mirza2014cgans]}. 
Traditional GANs are generative models that learn a mapping $G$ from a random noise
vector $z$ to an output image $y$, $G:z \to y$ \cite{[Goodfellow14]}.
Conditional GANs, instead, are conditioned to the input, in our case a periocular image taken in a specific spectrum. 
CGANs learn a mapping from an observed image $x$ and a random noise vector $z$ to the output $y$, $G:\left\{ {x,z} \right\} \to y$.
The objective of a Conditional GAN is usually expressed as:

\begin{equation}
\begin{aligned}
    \mathcal{L}_{cGAN}(G,D)= & \mathbb{E}_{x,y}[log D(x,y)] + \\ & \mathbb{E}_{x,y}[log (1 - D(x,G(x,z))]
\end{aligned}
\end{equation}

\noindent where $G$ (the generator) tries to minimize it against an
adversarial D (the discriminator) that tries to maximize it.

We make use of one of the most famous CGAN for image-to-image translation called Pix2Pix \cite{[isola16pix2pix]}. 
In contrast to other solutions like \cite{[Zhu17cycleGAN]}, where the network learns a mapping based on the general look of the input and output images, the method of \cite{[isola16pix2pix]} adds a L1 distance term to force the generated image to be closer to the target:


\begin{equation}
    \mathcal{L}_{L1}(G)=\mathbb{E}_{x,y,z}[\|y-G(x,z)\|_1]
\end{equation}

\noindent so that the loss function to optimize becomes:

\begin{equation}
\label{loss}
    G^{*}=arg \min_{G}\max_{D}\mathcal{L}_{cGAN}(G,D)+\lambda\mathcal{L}_{L1}(G)
\end{equation}

\noindent This way, 
the generator is
asked not to only fool the discriminator, but also to be near to the 
ground-truth image.


We use a Keras implementation of Pix2Pix \footnote{https://github.com/tjwei/GANotebooks}. 
The structure of the generator and discriminator is shown in Figure~\ref{fig:gan_pix2pix_models}.
The choice of this particular CGAN is because it exploits the pixel-wise alignment between images, as with the employed dataset. 
This should ease the work of the generator to get a closer representation of the output, and help to minimize the L1 term, giving a more accurate representation of the target data.
The generator uses a version of the famous encoder-decoder architecture U-NET \cite{ronneberger2015u}, in this case a U-NET-256 with batch normalization. 
The discriminator is a PatchGAN, where the output is an image of 30$\times$30, 
with each pixel corresponding 
to a patch of size 70$\times$70 in the input. 
This 70$\times$70 PatchGAN has proven to give sharper results both in spatial and spectral dimensions.
%
%
%
Finally, the discriminator checks if each patch belongs to the target image or has been generated, using the corresponding patch in the input image for comparison.
The model is trained using Adam as optimizer and a learning rate of 0.001.
%

\begin{figure}[t]
\centering
        \includegraphics[width=0.35\textwidth]{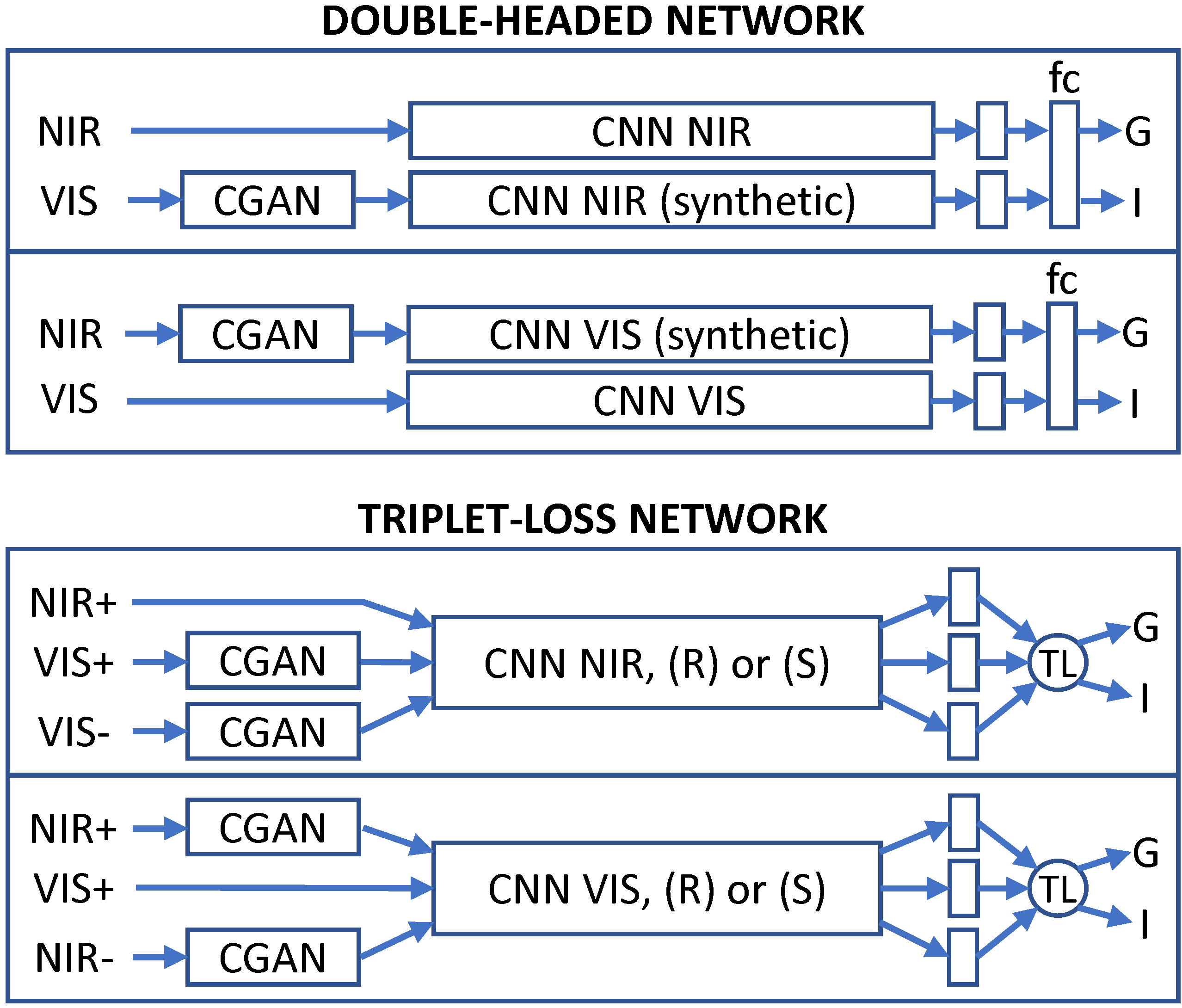}
\caption{Double-headed and triplet-loss networks for cross-spectral periocular recognition.}
\label{fig:CNNmodels}
\end{figure}


\begin{figure*}[htb]
\centering
        \includegraphics[width=0.72\textwidth]{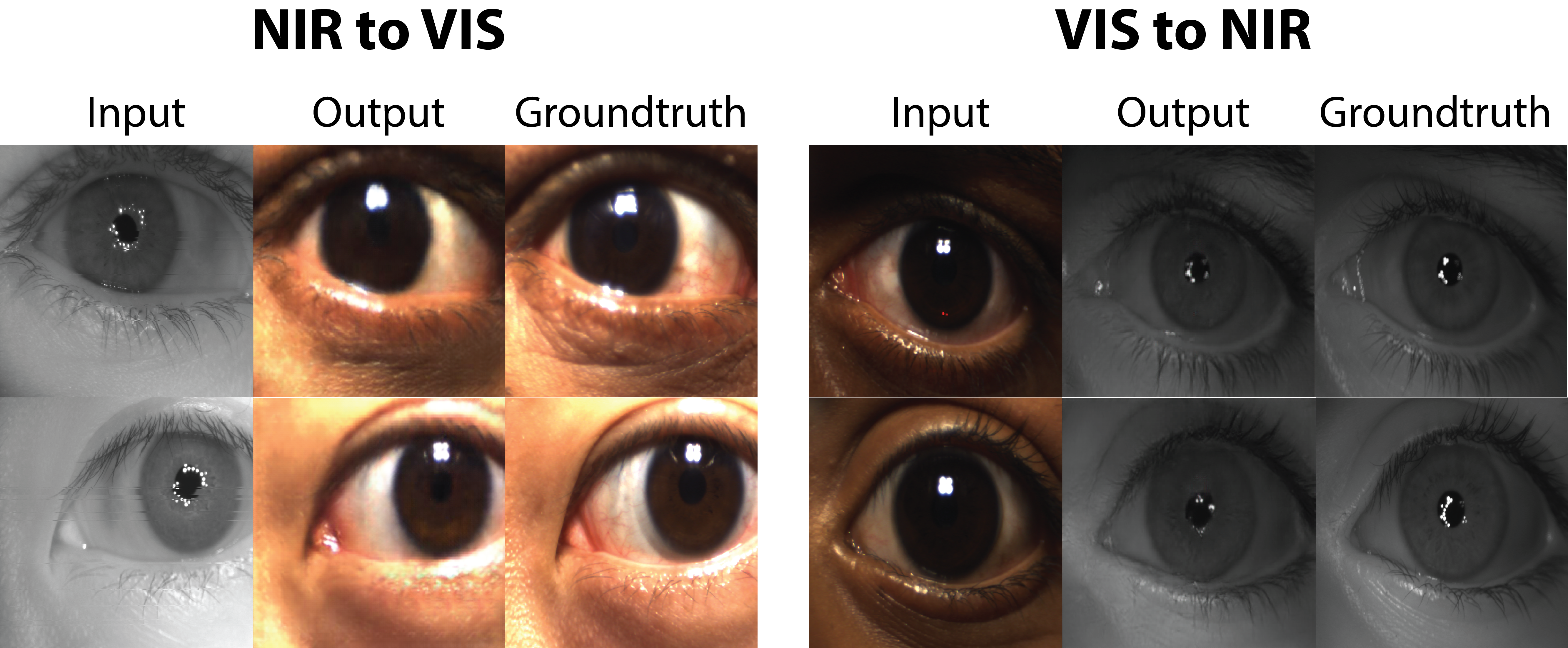}
\caption{Results of spectrum translation.}
\label{fig:results_spectrum_translation}
\end{figure*}

%
%
%
%
%
%
%
%

\begin{figure}[htb]
\centering
        \includegraphics[width=0.44\textwidth]{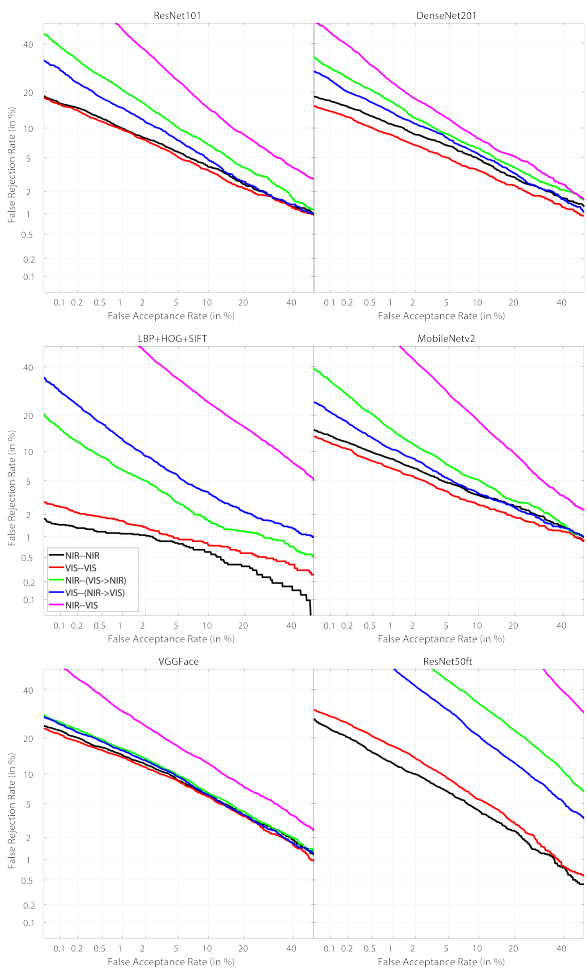}
\caption{Recognition performance using traditional periocular hand-crafted features and off-the-shelf CNNs. Best seen in color.}
\label{fig:results_OTS_features}
\end{figure}

\section{Experimental Framework}

\subsection{Database and Protocol}
\label{subsect:db_protocol}

We employ the Hong Kong Polytechnic University Cross-Spectral Iris Images Database (PolyU) \cite{[Nalla17]}.
It was acquired under simultaneous bi-spectral imaging in NIR and VIS wavelengths.
It contains 12540 iris images (209$\times$2$\times$15$\times$2)
from both eyes of 209 subjects, with 15 images per spectrum.
Each image is of 640$\times$480 pixels, with pixel correspondence between
the NIR and VIS images.
Some examples are shown in Figure~\ref{fig:sample_images_polyu}.
Following the procedure in \cite{[Nalla17]},
ten images from each eye are set aside as training images, 
while the remaining five images are used as test images.
Images are resized to 256$\times$256 pixels using bicubic interpolation to fit the generator requirement.

We carry out verification experiments, 
with each eye considered a different user,
leading to 209$\times$2=418 users.
Genuine trials are done by comparing each test image
of an eye to the remaining test images of the same eye,
avoiding symmetric comparisons. 
This results in 418$\times$(4+3+2+1)=4180 genuine scores.
Impostor trials are done by comparing the first
test image of an eye to the second test image of the remaining eyes,
resulting in 418$\times$417=174306 impostor scores.
Experiments have been done in a stationary computer with an i7-8700 processor, 32 Gb RAM, and a NVIDIA GTX Titan 5 GPU.
%

\subsection{Periocular Comparators}
\label{subsect:comparators}

To assess the goodness of the spectrum translation algorithm, we employ a number of periocular recognition experts, whose choice is motivated as follows. 
First, we apply the most widely used features in periocular research, employed as baseline in many studies
\cite{[Alonso16]}: Histogram of Oriented Gradients (HOG)
\cite{[Dalal05]}, Local Binary Patterns (LBP) \cite{[Ojala02]}, and
Scale-Invariant Feature Transform (SIFT) key-points \cite{[Lowe04]}.
The authentication for LBP and HOG consist of a simple Euclidean distance comparison between the feature images, while for SIFT we use the number of paired key-points normalized by the mean number of points in both images. We finally combine the scores of the three systems using Linear Logistic Regression, as explained in \cite{[Alonso20inffus]}.
Inspired by the works \cite{[Nguyen18],[Hernandez18],[Hernandez19]} in iris and ocular biometrics,
we also leverage the power of existing architectures pre-trained with
millions of images to classify hundreds of thousands of object categories\footnote{ImageNet. http://www.image-net.org}.
These have proven to
be successful in very large recognition tasks apart from the detection and
classification tasks for which they were designed \cite{[Razavian14]}.
Here, we employ the very deep
Resnet101 \cite{[He16]}
and Densenet201 \cite{[Huang17]}
architectures, as well as the mobile architecture MobileNet v2 \cite{[Sandler18mobilenetv2]}.
With this choice, we aim at comparing very deep architectures with a lightweight CNN.
In using these networks,
%
%
periocular images are fed into the feature extraction pipeline of each pre-trained CNN \cite{[Nguyen18],[Hernandez18]}.
But instead of using the vector from the last layer, we employ as feature descriptor
the vector from the intermediate layer identified as the one providing the best performance
(in our case, the layers recommended in \cite{[Alonso20inffus]} for NIR-VIS cross-spectral periocular recognition).
We also employ two networks pre-trained to classify faces: VGGFace \cite{[Parkhi15]}, and ResNet50ft \cite{[Cao18vggface2]}.
The first one is trained using the VGGFace dataset (2.6M images from 2.6K people), while the second one is trained using the MS-Celeb-1M dataset (10M images from 100K people) and then fine-tuned on VGGFace2 dataset (3.1M images from 8.6K people). Since these are trained
to classify faces, we speculate that they can provide effective recognition with the periocular
region as well, given that it appears in the training images.
These face pre-trained networks are also the models which are fine-tuned in \cite{[Zanlorensi20deepXspectralOcular]}.
With VGGFace, we use the layer recommended in \cite{[Alonso20inffus]}. The periocular descriptor with ResNet50ft is extracted from the layer adjacent to the classifier layer.
To carry out verification experiments, the extracted vectors with any of these networks can be simply compared using distance measures.
In our case, we employ the $\chi^2$ distance, which has shown better
results than other measures such as the cosine or Euclidean distances \cite{[Hernandez18]}.

%

To pursue ultimate performance, a ResNet50 \cite{[He16]} model, pretrained with the ImageNet dataset, has been fine-tuned to the periocular modality. This architecture has reported very good results in recent works applied to facial \cite{[Cao18vggface2]} and ocular recognition \cite{[Zanlorensi20deepXspectralOcular]}. 
Four networks are trained first to carry out biometric identification with the training images of the PolyU database: one network trained with NIR (real) images, one network with NIR (synthetic) images, one network with VIS (real) images, and finally, one network trained with VIS (synthetic) images.
Our motivation to train recognition networks with synthetic images is to assess the goodness of the spectrum translation algorithm by measuring differences in accuracy when trained with each type of data.
We then propose two different models for cross-spectral verification (Figure~\ref{fig:CNNmodels}). The first one consists of a double-headed CNN architecture, each branch being a previously trained CNN for identification with the specific type of data input (NIR or VIS, real or synthesized). We freeze the layers of the CNNs, and concatenate the outputs of the layers adjacent to the classifier layer to form an embedded vector from the two branches. Then, we train a fully connected neural network to classify the embedded vector as a genuine or an impostor attempt. We use Soft-Max as the last activation function, and train the system with binary cross-entropy loss and Adam as the optimizer.
The second system is a Triplet Loss network \cite{[Hoffer15triplet-loss-networks]}. We make use of our previously trained CNNs for identification as the base, with their layers frozen, and train a fully connected network to create an optimized small vector representation for verification. 
Here, the triplets are extracted with one CNN branch only, trained for the specific spectrum at hand.
In training the triplet-loss model, the back-bone network can be trained either with real images (`R' variant), or with synthetic images (`S' variant).
We use Euclidean distance, the improved triplet loss function \cite{[Zhang16improved-triplet-loss]} and Adam as the optimizer.

\begin{table*}[htb]
\scriptsize
\begin{center}
\begin{tabular}{ccccccc}

\multicolumn{7}{c}{} \\

Work & Classes & Modality & Features & EER & GAR@FAR=1\%  & GAR@FAR=0.1\% \\ \hhline{=======}  

\cite{[Behera17]} & 209 &  Periocular (L)  & LBP & 36.61\% & - & 25.56\% \\ \cline{3-7}
                  & &  Periocular (L) & HOG & 19.57\% & - & 70.97\% \\ \cline{3-7}
                  
                  & &  Periocular (R) & LBP & 35.74\% & - & 29.43\% \\ \cline{3-7}
                  & &  Periocular (R) & HOG & 18.79\% & - & 73.12\% \\ \cline{3-7}
                  
                  & &  Periocular (L+R) & LBP & 35.55\% & - & 35.37\% \\ \cline{3-7}
                  & &  Periocular (L+R) & HOG & 13.87\% & - & 83.12\% \\ \hhline{=======} 

\cite{[Ramaiah16]} & 280 &  Iris & LG & 33.4\% & - & 41.8\% \\ \cline{3-7}
                   & &  Iris & MRF & - & - & 61.9\% \\ \cline{3-7}
                   & &  Periocular & FPLBP & 32.5\% & - & 45.4\% \\ \cline{3-7}
                   & &  Periocular & TPLBP & 19.8\% & - & 73.2\% \\ \cline{3-7}
                   & &  Fusion & LG+FPLBP & 26.6\% & - & - \\ \cline{3-7}
                   & &  Fusion & LG+TPLBP & 17.9\% & - & 74.7\% \\ \cline{3-7}
                   & &  Fusion & LG+TPLBP+MRF & - & - & 84.2\% \\ \hhline{=======} 
                   
\cite{[Nalla17]} & 280 &  Iris & NNBN & 26.68\% & - & 58.8\% \\ \cline{3-7}
                 & &  Iris & MRF & 23.87\% & - & 61.9\% \\ \hhline{=======}  
                 
\cite{[Wang19]} & 280 &  Iris & CNN-SDH & 5.39\% & 90.7\% & 86.6\% \\ \cline{3-7}
                &  &  Iris & VGG16-SDH & 4.85\% & 91.6\% & 83.2\% \\ \cline{3-7}
                &  &  Iris & ResNet50-SDH & 7.17\% & 87.9\% & 76.8\% \\ \hhline{=======}  

\multicolumn{7}{c}{} \\ 

Work & Classes & Modality & Features & EER & GAR@FAR=1\%  & GAR@FAR=0.1\% \\ \hhline{=======} 

\cite{[Wang19]} & 418 & Iris & CNN-SDH & 12.41\% & 64.2\% & 57.4\% \\ \hhline{=======} 

\cite{[Zanlorensi20deepXspectralOcular]} & 418 & Iris & VGGFace & 2.16$\pm$0.16\% & - & - \\ \cline{3-7}
                                         &     & Iris & ResNet50ft & 1.13$\pm$0.14\% & $\sim$98.8\% & $\sim$94.7\% \\  \cline{3-7}
                                         
                                         &     & Periocular & VGGFace & 1.8$\pm$0.21\% & -& - \\ \cline{3-7}
                                         &     & Periocular & ResNet50ft & \textbf{0.78$\pm$0.09\%} & \textbf{$\sim$99.3\%} & \textbf{$\sim$98.2\%} \\ \cline{3-7}           
                                         &     & Fusion & VGGFace & 0.93$\pm$0.1\% & - & - \\ \cline{3-7}
                                         &     & Fusion & ResNet50ft & 0.49$\pm$0.06\% & $\sim$99.6\% & $\sim$99.2\% \\ \hhline{=======}

Ours & 418 &  Periocular        &  ResNet101  &   12.6\% &  53.5\% &  26.7\%  \\  \cline{4-7}
     &     &  NIR-VIS           &  DenseNet201  &   8.6\% &  76.9\% &  54.7\%  \\  \cline{4-7}
     &     &                    &  MobileNetv2  &   13.6\% &  46.6\% &  20.7\%  \\  \cline{4-7}
     &     &                    &  LBP+HOG+SIFT  &   17.5\% &  43.8\% &  20.4\%  \\  \cline{4-7}
     &     &                    &  VGGFace  &   11.4\% &  69.9\% &  48.4\%  \\  \cline{4-7}
     &     &                    &  ResNet50ft  &   39.6\% &  3.9\% &  1.1\%  \\  \hhline{~~~====}
     &     &                    &  ResNet50+SM  &  1.9\% &  96.5\% &  67.6\%  \\  \cline{3-7}
                
     &     &  Periocular        &  ResNet101  &   8.0\% &  79.4\% &  61.4\%  \\  \cline{4-7}
     &     &  NIR-(VIS$\to$NIR)  &  DenseNet201  &   7.2\% &  83.5\% &  70.7\%  \\  \cline{4-7}
     &     &                    &  MobileNetv2  &   6.4\% &  84.8\% &  66.6\%  \\  \cline{4-7}
     &     &                    &  LBP+HOG+SIFT  &   3.6\% &  93.5\% &  84.6\%  \\  \cline{4-7}
     &     &                    &  VGGFace  &   7.6\% & 83.5\% &  74.3\%  \\  \cline{4-7}
     &     &                    &  ResNet50ft  &   21.5\% &  36.4\% &  16.7\%  \\  \hhline{~~~====}
     &     &                    &  ResNet50+SM  &   1.5\% &  97.4\% &  74.6\%  \\  \cline{4-7}
     &     &                    &  ResNet50(R)+TL  &   1.5\% &  98.0\% &  93.0\%  \\\cline{4-7}
     &     &                    &  ResNet50(S)+TL  &   \textbf{1.0\%} &  \textbf{99.1\%} &  \textbf{96.7\%}  \\  \cline{3-7}
     &     &  Periocular        &  ResNet101  &   6.5\% &  85.1\% &  71.8\%  \\  \cline{4-7}
     &     &  VIS-(NIR$\to$VIS)  &  DenseNet201  &   6.7\% &  86.4\% &  75.7\%  \\  \cline{4-7}
     &     &                    &  MobileNetv2  &   5.1\% & 89.7\% &  79.0\%  \\  \cline{4-7}
     &     &                    &  LBP+HOG+SIFT  &   5.3\% &  87.5\% &  71.0\%  \\  \cline{4-7}
     &     &                    &  VGGFace  &   7.3\% &  84.0\% &  75.0\%  \\  \cline{4-7}
     &     &                    &  ResNet50ft  &   15.3\% &  49.7\% &  27.9\%  \\  \hhline{~~~====}
     &     &                    &  ResNet50+SM  &   2.0\% &  96.3\% &  77.5\% \\  \cline{4-7}
     &     &                    &  ResNet50(R)+TL  &   1.7\% &  97.6\% &  91.3\%  \\ \cline{4-7}
     &     &                    &  ResNet50(S)+TL  &   1.9\% &  97.4\% &  93.0\%  \\  \hhline{=======}

\end{tabular}

\end{center}
\caption{Cross-spectral performance of other works in the literature using the same database. The best results of the periocular modality are marked in bold in each column, as well as the best results achieved with our approach.}
\label{tab:otherworks-results}
\end{table*}
\normalsize

\subsection{Results and Discussion}
\label{subsect:results}

Figure~\ref{fig:results_OTS_features} presents the results of our experiments with the hand-crafted features and off-the-shelf CNN descriptors described in Section~\ref{subsect:comparators}.
We report both intra-spectral (black/red curves) and cross-spectral experiments. 
Cross-spectral experiments include matching of features from image pairs in different spectra (NIR-VIS, purple curves), and matching of features from pairs where one of the images has been translated to the spectrum of the other image with the CGAN method of Section~\ref{sect:spectrum_translation}. This involves two possibilities: that the images are compared in the NIR spectrum (indicated as NIR-(VIS$\to$NIR), green curves), and that the images are compared in the VIS spectrum (indicated as VIS-(NIR$\to$VIS), blue curves).
Numerical results of cross-spectral experiments with these off-the-shelf descriptors are given in Table~\ref{tab:otherworks-results} (bottom) as well.
Some example images after applying spectrum translation 
are also shown in Figure~\ref{fig:results_spectrum_translation},
together with the input and target (ground-truth) images,
where we can appreciate the realistic results obtained.

\begin{figure}[htb]
\centering
        \includegraphics[width=0.42\textwidth]{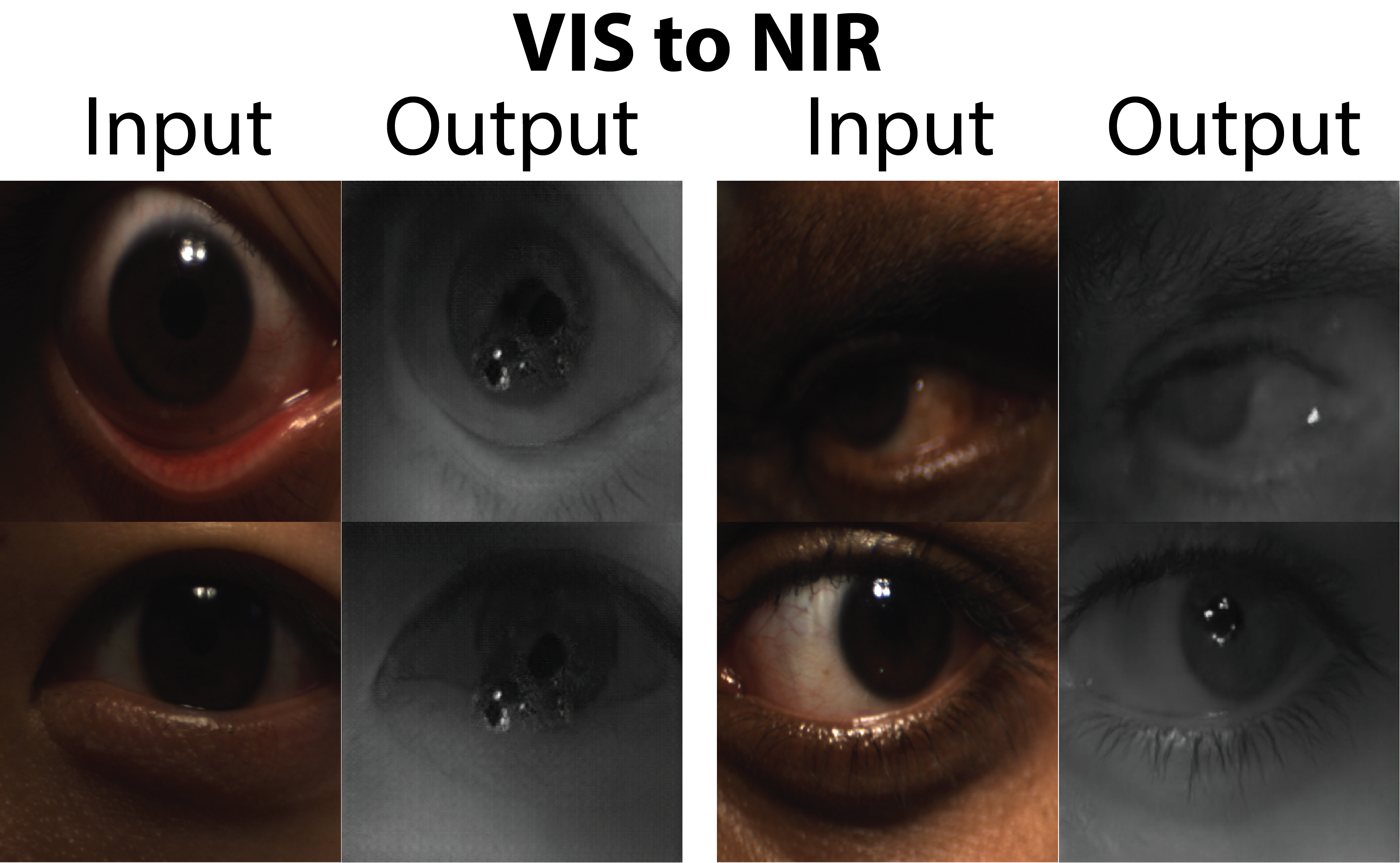}
\caption{Results of (not so good) spectrum translation.}
\label{fig:results_spectrum_translation_fail}
\end{figure}

From Figure~\ref{fig:results_OTS_features}, we observe that translating images to the same spectrum provides improved performance (green/blue curves), in comparison to keeping images in their original spectra (purple curves). 
The purple curves have an EER between 8.6 and 17.5\%, while the EER after translating both images to the same spectrum is between 3.6 and 8\% (if we do not consider the ResNet50ft descriptor, which has much worse performance than the others).
In some cases (VGGFace), cross-spectral performance is even comparable to the intra-spectral case. This validates our approach of converting images to the same spectrum before verification, instead of matching features extracted from images in different spectra. 
It is also worth noting that the majority of systems exhibit better performance if images are converted to the VIS spectrum (blue vs. green curve). 
A possible explanation for this can be because of the
creation of noisy iris textures in the generator when
synthesizing NIR images, 
together with the fact that skin texture is less visible in
the NIR spectrum (see Figure~\ref{fig:results_spectrum_translation}).
On the other hand,
VIS images given by the generator
seem to recreate more similar skin color and textures to the original image.
In Figure~\ref{fig:results_spectrum_translation_fail}, we also give some
examples where the spectrum translation algorithm struggles,
showing the mentioned phenomenon of creation of noisy iris textures
when generating NIR images. 
In the upper images, the pupil appears with irregular shape and disseminated. In the lower images, the generated pupil does not appear or is not aligned with the iris. The latter seems to be caused due to the eye not being looking to the front in those particular images, while the majority of eyes in the dataset are gazing to the camera, so the generator does not see sufficient training images having this phenomena.
%
Overall it also looks like the translation gives worse performance more under low illumination.

By looking at the performance of the individual systems, it is surprising that the combination of the hand-crafted HBP+HOG+SIFT features stands out as a top performer, surpassing deep architectures such as DenseNet201 or ResNet101. 
It is also remarkable the good performance of MobileNet v2, a mobile model with just 53 layers, in comparison to ResNet101 (101 layers) or DenseNet201 (201 layers).
Also worth noting, the models pre-trained with face images (VGGFace and ResNet50ft) do not show better performance than the others, despite they have seen images containing the periocular region during training. 
In particular, the performance of ResNet50ft is surprisingly bad in comparison with the other descriptors.
On the other hand, VGGFace shows comparable performance between the intra-spectral and cross-spectral scenarios (after image translation), although its intra-spectral performance is already worse than the other descriptors shown in Figure~\ref{fig:results_OTS_features}.

We also compare the performance of our approach with another works in the literature 
that employ the same database. Results are given in Table~\ref{tab:otherworks-results}, 
including the fine-tuned ResNet50 architecture described in Section~\ref{subsect:comparators}.
Results show the two models proposed for verification (Figure~\ref{fig:CNNmodels}): 
the double-headed CNN with Soft-Max activation 
(ResNet50+SM), and the Triplet-Loss network (ResNet50+TL), the latter including the two variants for the back-bone network (`R' and `S').
%
%
Some works reported in Table~\ref{tab:otherworks-results} are focused on the periocular modality \cite{[Behera17]}, while
another works employ the iris region only, or a fusion of both modalities \cite{[Ramaiah16],[Nalla17],[Wang19],[Zanlorensi20deepXspectralOcular]}. 
It should also be noted that the number of classes employed is different.
For example, in \cite{[Behera17]} they use 209 classes (corresponding to the 209 individuals of the database),
reporting results for each eye separately, and for their feature-level combination.
As expected, combining features from both left and right eyes leads to a better performance.
The works \cite{[Ramaiah16],[Nalla17],[Wang19]} employ a subset of 280 eyes 
from the 209$\times$2=418 eyes available in the database. They correspond 
to 140 subjects whose images are properly segmented in both VIS and NIR channels
using an automatic segmentation algorithm. 
From this viewpoint, the images employed are those with sufficient quality so that an automatic algorithm is able to locate the iris region successfully.
%
%
In \cite{[Wang19],[Zanlorensi20deepXspectralOcular]}, they also report results using the entire set of 418 eyes available, each eye being a different class,
which is the same setup employed in this paper.
The difference between using 280 and 418 classes is also assessed in \cite{[Wang19]}, with the performance comparatively degraded from EER=5.39\% to 12.41\%.

As it can be observed, our approach outperform all previous studies (except the work \cite{[Zanlorensi20deepXspectralOcular]}), even if we are using the entire dataset 
instead of a sub-set of high quality images.
Considering that we are using the entire database with low quality images included,
it is also worth noting that
the cross-sensor EER obtained with some off-the-shelf descriptors like
LBP+HOG+SIFT or MobileNetv2 (around 4-5\%) 
would enable low security applications 
without the need of training a dedicated verification network.
At FAR=1\%, their performance is still acceptable (GAR of 89.7\% or higher), 
enabling some applications requiring some level of security as well.
This validates the suggested approach of converting the images to the same spectrum for comparison.

Regarding the work \cite{[Zanlorensi20deepXspectralOcular]}, it employs 
the VGGFace and ResNet50ft networks pre-trained for face recognition (similar to our off-the-shelf networks)
\cite{[Parkhi15],[Cao18vggface2]},
which are then fine-tuned for periocular recognition with the PolyU database.
%
In particular, the pre-training of ResNet50ft consists of 
a double fine-tuning with more than 13M face images from nearly 110K people (as described in \cite{[Cao18vggface2]}). 
In contrast, our ResNet50 model is pre-trained with a database comparatively much smaller (ImageNet, with about 1M images from 1000 classes) before it is fine-tuned with PolyU.
Still, our best configuration achieves an EER of just 0.22\% above the experiments of \cite{[Zanlorensi20deepXspectralOcular]}, with a much less training budget.
One behaviour that we can appreciate also is that, even though the Soft-Max (SM) and the Triplet Loss (TL) approaches give similar EERs, the latter seems to be more stable under a more restrictive low FAR. 
One possible explanation could be the exponential behavior of the soft-max function. Since triplet loss networks use distance metrics, they show a more progressive degradation. 
Another interesting results are that 
$i$) the best performance with our trained models is obtained if both images are compared in the NIR spectrum, in contrast to the off-the-shelf descriptors, which showed better performance in the VIS spectrum; and
$ii$) a better performance is obtained if the back-bone network of the triplet loss approach is trained with synthetic (S) NIR images.
The latter opens the door to combine real and synthetic images to augment the training dataset.

Finally, in using the PolyU database, the work \cite{[Wang19]} trains a CNN architecture similar to AlexNet, which is shallower in comparison to more recent deep architectures such as VGGFace or ResNet50 employed in \cite{[Zanlorensi20deepXspectralOcular]} and in the present work. 
This could be an explanation of the better performance shown by more recent studies with the entire PolyU database.
The deeper architecture of ResNet50 in particular has demonstrated excellent results when fine-tuned to another modalities such as face \cite{[Cao18vggface2]}, and it also shows superior capabilities in the periocular modality, as reported here.
Also, although the work \cite{[Wang19]} employs the iris modality, 
it is worth to mention that our experiments with the periocular modality outperforms the results of \cite{[Wang19]} by a large margin after using the proposed spectrum translation technique. 
The superiority of our approach is observed even with off-the-shelf descriptors which are not trained specifically for the periocular modality.
Even without translating images to the same spectrum (NIR-VIS case), some off-the-shelf descriptors employing very deep networks already provide a better performance as well, e.g. DenseNet201 (EER=8.6\% vs. 12.41\%) or VGGFace (EER=11.4\%). 
The superiority of the periocular modality for cross-spectral recognition in the difficult imaging conditions of the PolyU database has been also observed in other works that compare both modalities, e.g. \cite{[Ramaiah16],[Zanlorensi20deepXspectralOcular]}.
Apart than the iris constitutes a smaller region of the image,
another explanation could be that iris recognition is more sensitive to cross-spectral experiments due to melanin in dark irises (preponderant in this database), which causes significant differences in how the iris region appears in each spectrum. 
In this sense, it would be interested to test if the spectrum translation method proposed is able to reduce such differences, helping to close the gap between the two biometric modalities.

%
%

\section{Conclusions}

This work investigates the challenge of comparing periocular images captured in different spectra, which usually provides worse performance
than if they were captured in the same spectrum \cite{[Jillela14],[Sharma14],[Ramaiah16],[Hernandez19]}.
We propose the use of an image translation technique based on 
Conditional Generative Adversarial Networks (CGANs) \cite{[isola16pix2pix]}.
The algorithm employed is trained to translate images from the visible spectrum (VIS)
to the near-infrared spectrum (NIR), or viceversa, so that biometric verification is
done in the same spectrum. 
The proposed method has the advantage that traditional feature methods 
trained to work in a single spectrum can be used out-of-the-box,
with a number of different approaches evaluated in this paper.
Experimental results are given on a database of 12540 images 
in the VIS and NIR spectra from both eyes of 209 persons
(giving 209$\times$2=418 available eyes). 
Our experiments show the superiority
of the proposed approach in comparison to previous studies that
employ the same database.
To further reduce the error rates,
we also fine-tune two different systems for cross-spectral biometric verification 
that make use of ResNet50 
pre-trained on ImageNet as the base architecture (Figure~\ref{fig:CNNmodels}). 
The first one employs a double-headed network trained with Soft-Max and Cross-Entropy loss, 
and the second one is Triplet Loss network trained to create embedding of genuine vectors 
that are close to each other while pushing impostor vectors apart. 
A particularity is that the CNN weights used for feature extraction 
are trained previously for biometric identification in a single spectrum, 
and then the networks are frozen to train the Soft-Max or Triplet Loss approach.
This helps to reduce computation time and resources.
Also, the back-bone networks operate in a single spectrum, 
so images are previously converted using the proposed spectrum translation technique.
With this approach, we obtain results competitive with the state-of-the-art \cite{[Zanlorensi20deepXspectralOcular]}, but with a
more reduced training scheme, achieving a cross-spectral
periocular performance of EER=1\% and GAR=99.1\% @ FAR=1\%.
We also outperform previously reported results, even if operating in less-favorable conditions,
i.e. by maintaining low-quality images that are removed in other studies in the literature.
This is even achieved with off-the-shelf general descriptors that are not specifically trained
for the periocular modality, validating the employed approach of
converting the images to the same spectrum before comparison.

%
%
%

Future work includes the exploration of image translation approaches
that do not need pixel-to-pixel correspondence for training, 
such as CycleGAN \cite{[Zhu17cycleGAN]}.
We will also explore the inclusion of a biometric recognition term
to the generator loss function.
Given the success in employing synthetic images to train the verification networks,
augmenting the training set with inclusion of both real and synthetic images will be another avenue.
We are also interested in evaluating the verification system when the layers of the
CNNs are not frozen, and in the applicability of our cross-spectral recognition
to other modalities such as iris or face.

\section*{Acknowledgements}

The authors thank the Swedish Research Council for funding
their research. Part of the computations were enabled by resources provided by the Swedish National Infrastructure for Computing (SNIC) at NSC Linköping. We also gratefully acknowledge the support of NVIDIA with the donation of the Titan V GPU used for this research.

{\small
\bibliographystyle{ieee}

\begin{thebibliography}{10}\itemsep=-1pt

\bibitem{[Wang19]}
Cross-spectral iris recognition using cnn and supervised discrete hashing.
\newblock {\em Pattern Recognition}, 86:85 -- 98, 2019.

\bibitem{[Alonso16]}
F.~Alonso-Fernandez, J.~Bigun.
\newblock A survey on periocular biometrics research.
\newblock {\em Pattern Recogn. Letters}, 82:92--105, 2016.

\bibitem{[Alonso18_perioc_expression]}
F.~{Alonso-Fernandez}, J.~{Bigun}, and C.~{Englund}.
\newblock Expression recognition using the periocular region: A feasibility
  study.
\newblock In {\em 14th Intl Conf on Signal-Image Technology
  Internet-Based Systems (SITIS)}, pages 536--541, 2018.

\bibitem{[Alonso20inffus]}
F.~Alonso{-}Fernandez, K.~B. Raja, R.~Raghavendra, C.~Busch, J.~Big{\"{u}}n,
  R.~Vera{-}Rodr{\'{\i}}guez, and J.~Fi{\'{e}}rrez.
\newblock Cross-sensor periocular biometrics: {A} comparative benchmark
  including smartphone authentication.
\newblock {\em CoRR}, abs/1902.08123, 2019.

\bibitem{[Behera17]}
S.~S. Behera, M.~Gour, V.~Kanhangad, and N.~Puhan.
\newblock Periocular recognition in cross-spectral scenario.
\newblock In {\em Proc IEEE Intl Joint Conf on Biometrics,
  IJCB}, pages 681--687, 2017.

\bibitem{[Cao18vggface2]}
Q.~{Cao}, L.~{Shen}, W.~{Xie}, O.~M. {Parkhi}, A.~{Zisserman}.
\newblock Vggface2: A dataset for recognising faces across pose and age.
\newblock In {\em 13th IEEE Intl Conf on Automatic Face and
  Gesture Recognition (FG 2018)}, pages 67--74, 2018.

\bibitem{[Dalal05]}
N.~{Dalal}, B.~{Triggs}.
\newblock Histograms of oriented gradients for human detection.
\newblock In {\em Proc IEEE Conf on Computer Vision and
  Pattern Recognition (CVPR)}, vol~1, pages 886--893, 2005.

\bibitem{[Goodfellow14]}
I.~Goodfellow, J.~Pouget-Abadie, M.~Mirza, B.~Xu, D.~Warde-Farley, S.~Ozair,
  A.~Courville, and Y.~Bengio.
\newblock Generative adversarial nets.
\newblock In Z.~Ghahramani, M.~Welling, C.~Cortes, N.~D. Lawrence, and K.~Q.
  Weinberger, editors, {\em Advances in Neural Information Processing Systems
  27}, pages 2672--2680. Curran Associates, Inc., 2014.

\bibitem{[He16]}
K.~He, X.~Zhang, S.~Ren, J.~Sun.
\newblock Deep residual learning for image recognition.
\newblock In {\em Proc IEEE Conf on Computer Vision and Pattern
  Recognition, CVPR}, pages 770--778, June 2016.

\bibitem{[Hernandez18]}
K.~Hernandez-Diaz, F.~Alonso-Fernandez, J.~Bigun.
\newblock Periocular recognition using {CNN} features off-the-shelf.
\newblock In {\em Proc Intl Conf of the Biometrics Special
  Interest Group (BIOSIG)}, pages 1--5, Sep. 2018.

\bibitem{[Hernandez19]}
K.~Hernandez-Diaz, F.~Alonso-Fernandez, and J.~Bigun.
\newblock Cross spectral periocular matching using resnet features.
\newblock In {\em Proc International Conference on Biometrics (ICB)}, 2019.

\bibitem{[Hoffer15triplet-loss-networks]}
E.~Hoffer and N.~Ailon.
\newblock Deep metric learning using triplet network.
\newblock In {\em International Workshop on Similarity-Based Pattern
  Recognition}, pages 84--92. Springer, 2015.

\bibitem{[Huang17]}
G.~{Huang}, Z.~{Liu}, L.~v.~d. {Maaten}, and K.~Q. {Weinberger}.
\newblock Densely connected convolutional networks.
\newblock In {\em 2017 IEEE Conference on Computer Vision and Pattern
  Recognition (CVPR)}, pages 2261--2269, 2017.

\bibitem{[isola16pix2pix]}
P.~Isola, J.-Y. Zhu, T.~Zhou, and A.~A. Efros.
\newblock Image-to-image translation with conditional adversarial networks,
  2016.

\bibitem{[Jillela14]}
R.~R. Jillela and A.~Ross.
\newblock Matching face against iris images using periocular information.
\newblock {\em Proc Intl Conf Image Processing, ICIP}, pages 4997--5001, Oct
  2014.

\bibitem{[Juefei-Xu11]}
F.~Juefei-Xu, K.~Luu, M.~Savvides, T.~Bui, and C.~Suen.
\newblock Investigating age invariant face recognition based on periocular
  biometrics.
\newblock {\em Proc Intl Joint Conf Biometrics, IJCB}, Oct 2011.

\bibitem{[Lowe04]}
D.~Lowe.
\newblock Distinctive image features from scale-invariant key points.
\newblock {\em Intl Journal of Computer Vision}, 60(2):91--110, 2004.

\bibitem{[Miller10]}
P.~E. {Miller}, J.~R. {Lyle}, S.~J. {Pundlik}, and D.~L. {Woodard}.
\newblock Performance evaluation of local appearance based periocular
  recognition.
\newblock In {\em Proc Fourth IEEE Intl Conf on Biometrics: Theory,
  Applications and Systems (BTAS)}, pages 1--6, 2010.

\bibitem{[mirza2014cgans]}
M.~Mirza and S.~Osindero.
\newblock Conditional generative adversarial nets.
\newblock {\em arXiv preprint arXiv:1411.1784}, 2014.

\bibitem{[Nalla17]}
P.~R. {Nalla}, A.~{Kumar}.
\newblock Toward more accurate iris recognition using cross-spectral matching.
\newblock {\em IEEE TIP}, 26(1), 2017.

\bibitem{[Nguyen18]}
K.~Nguyen, C.~Fookes, A.~Ross, and S.~Sridharan.
\newblock Iris recognition with off-the-shelf cnn features: A deep learning
  perspective.
\newblock {\em IEEE Access}, 6:18848--18855, 2018.

\bibitem{[Nigam15]}
I.~Nigam, M.~Vatsa, and R.~Singh.
\newblock Ocular biometrics: A survey of modalities and fusion approaches.
\newblock {\em Information Fusion}, 26:1 -- 35, 2015.

\bibitem{[Ojala02]}
T.~Ojala, M.~Pietikainen, T.~Maenpaa.
\newblock Multiresolution gray-scale and rotation invariant texture
  classification with local binary patterns.
\newblock {\em IEEE TPAMI},
  24(7):971--987, 2002.

\bibitem{[Park11]}
U.~Park, R.~R. Jillela, A.~Ross, A.~K. Jain.
\newblock Periocular biometrics in the visible spectrum.
\newblock {\em IEEE TIFS}, 6(1),
  2011.

\bibitem{[Parkhi15]}
O.~M. Parkhi, A.~Vedaldi, and A.~Zisserman.
\newblock Deep face recognition.
\newblock In M.~W.~J. Xianghua~Xie and G.~K.~L. Tam, editors, {\em Proc. British Machine Vision Conference (BMVC)}, pages 41.1--41.12. BMVA
  Press, September 2015.

\bibitem{[Ramaiah16]}
N.~P. Ramaiah, A.~Kumar.
\newblock On matching cross-spectral periocular images for accurate biometrics
  identification.
\newblock In {\em Proc BTAS}, pages 1--6, 2016.

\bibitem{[Razavian14]}
A.~S. Razavian, H.~Azizpour, J.~Sullivan, S.~Carlsson.
\newblock Cnn features off-the-shelf: An astounding baseline for recognition.
\newblock In {\em Proc IEEE Conference on Computer Vision and Pattern
  Recognition Workshops, CVPRW}, pages 512--519, 2014.

\bibitem{ronneberger2015u}
O.~Ronneberger, P.~Fischer, T.~Brox.
\newblock U-net: Convolutional networks for biomedical image segmentation.
\newblock In {\em Intl Conf on Medical image computing and
  computer-assisted intervention}, pages 234--241. Springer, 2015.

\bibitem{[Sandler18mobilenetv2]}
M.~{Sandler}, A.~{Howard}, M.~{Zhu}, A.~{Zhmoginov}, and L.~{Chen}.
\newblock Mobilenetv2: Inverted residuals and linear bottlenecks.
\newblock In {\em IEEE/CVF Conference on Computer Vision and Pattern
  Recognition, CVPR}, pages 4510--4520, 2018.

\bibitem{[sequeira16crosseyed]}
A.~F. Sequeira, L.~Chen, J.~Ferryman, F.~Alonso-Fernandez, J.~Bigun, K.~B.
  Raja, R.~Raghavendra, C.~Busch, and P.~Wild.
\newblock Cross-eyed - cross-spectral iris/periocular recognition database and
  competition.
\newblock In {\em Proc Intl Conf of the Biometrics Special Interest Group,
  BIOSIG}, pages 1--5, 2016.

\bibitem{[Sharma14]}
A.~{Sharma}, S.~{Verma}, M.~{Vatsa}, R.~{Singh}.
\newblock On cross spectral periocular recognition.
\newblock In {\em Proc IEEE International Conference on Image Processing
  (ICIP)}, pages 5007--5011, 2014.

\bibitem{[Zanlorensi20deepXspectralOcular]}
L.~A. {Zanlorensi}, D.~R. {Lucio}, A.~d.~S.~{Britto Junior}, H.~{Proença}, D.~{Menotti}.
\newblock Deep representations for cross-spectral ocular biometrics.
\newblock {\em IET Biometrics}, 9(2), 2020.

\bibitem{[Zhang16improved-triplet-loss]}
S.~Zhang, Y.~Gong, J.~Wang.
\newblock Deep metric learning with improved triplet loss for face clustering
  in videos.
\newblock In {\em Pacific Rim Conference on Multimedia}, pages 497--508.
  Springer, 2016.

\bibitem{[Zhu17cycleGAN]}
J.~{Zhu}, T.~{Park}, P.~{Isola}, A.~A. {Efros}.
\newblock Unpaired image-to-image translation using cycle-consistent
  adversarial networks.
\newblock In {\em 2017 IEEE International Conference on Computer Vision
  (ICCV)}, pages 2242--2251, 2017.

\end{thebibliography}

}

\end{document}